\documentclass{article}

\PassOptionsToPackage{numbers, compress}{natbib}

\usepackage[main,final]{neurips_2026}

\usepackage[utf8]{inputenc} 
\usepackage[T1]{fontenc}    

\usepackage[pagebackref,breaklinks,colorlinks,allcolors=cyan]{hyperref}
\usepackage{booktabs}       
\usepackage{amsfonts}       
\usepackage{nicefrac}       
\usepackage{microtype}      
\usepackage{xcolor}         
\usepackage{amsmath}
\usepackage{amsthm}
\newtheorem{proposition}{Proposition}
\newtheorem{theorem}{Theorem}
\newtheorem{lemma}{Lemma}
\newtheorem{assumption}{Assumption}
\usepackage{multirow}
\usepackage{graphicx}
\usepackage{subfig}
\usepackage{adjustbox} %
\usepackage{comment}

\renewcommand{\acksection}{\section*{Acknowledgments}}
\title{RankAlign: Vision-Language Representation Alignment via Rank Transformation}

\author{%
  Enzhe Zhao\textsuperscript{1,2}\thanks{Equal contribution. Correspondence to: \texttt{marco.fiorucci@unipd.it}.}\quad
  \textbf{Marco Fiorucci\textsuperscript{1}\footnotemark[1]}\quad
  \textbf{Lamberto Ballan\textsuperscript{1}} \vspace{1em} \\
  \textsuperscript{1}~University of Padova, Italy \vspace{.1em}\\ \textsuperscript{2}~Harbin Institute of Technology, China \vspace{0.5em} \\
    {{\small \textbf{Project page:} {\tt\href{https://marcofiorucci.github.io/projects/rankalign/}{marcofiorucci.github.io/rankalign/}} }%
    }}

\begin{document}

\maketitle

\begin{abstract}
While the Platonic Representation Hypothesis posits that vision and language models converge toward compatible latent geometries, recovering class-level correspondences between unpaired modalities without cross-modal supervision remains challenging.
Existing relational matching methods compare within-modality similarity kernels directly, although the same semantic structure may be represented through different numerical parameterizations across modalities. We propose RankAlign, a rank-based normalization strategy for class-level unpaired vision-language matching. RankAlign transforms each within-modality similarity kernel into an ordinal relational representation before permutation matching, while
leaving the correspondence space and matching solver unchanged. The resulting representation is invariant to strictly increasing reparameterizations of similarity values while preserving the ordinal structure used for matching. We provide exact-recovery and stability guarantees that characterize correspondence recovery under both exact and partial ordinal agreement. Experiments across diverse frozen vision and language encoders and multiple datasets demonstrate improved and more stable class-level correspondence recovery, including under label corruption and domain shift. 
\end{abstract}

\section{Introduction}
Recent advances in foundation models have fundamentally redefined representation learning across diverse modalities. While self-supervised vision encoders like DINO~\cite{caron2021dino} and large-scale language models \cite{devlin2019bert, reimers2019sentence} have matured within their respective domains, multimodal contrastive frameworks such as CLIP~\cite{radford2021clip} and SigLIP2~\cite{tschannen2025siglip} have further bridged the two modalities. These architectures induce sophisticated latent representations that capture intricate structural and semantic nuances. Yet, despite this progress, the geometric alignment between visual and language representation spaces remains poorly understood when encoders are trained independently on disparate datasets. This poses a significant challenge for class-level unpaired cross-modal matching, where category structure is available within each modality, while the correspondence between visual and textual categories remains unknown and must be recovered.

\textbf{Representation similarity and the Platonic hypothesis.} Independently trained visual and language models frequently exhibit a striking convergence toward compatible representational structures. This phenomenon is formalized by the Platonic Representation Hypothesis, which posits that scaling and data diversity drive neural networks toward shared latent geometries that encode the statistical regularities of the world~\cite{huh2024position}. In vision-language settings, visual and textual encoders learn compatible relational structures despite disparate training objectives~\cite{maniparambil2024world, schnaus2025blind}. Although methods such as CCA~\cite{cca1936}, SVCCA~\cite{raghu2017svcca}, and CKA~\cite{kornblith2019similarity} provide quantitative measures of representational similarity, they rely on direct numerical comparisons that are inherently sensitive to modality-specific scale distortions and biases.

\textbf{Class-level unpaired vision-language matching.}
The emergence of a shared embedding space raises the fundamental question: can explicit correspondences between vision and language representations be recovered when independently trained encoders receive no paired supervision?
Traditional alignment relies on image--caption pairs or manually verified correspondences to learn a cross-modal mapping.
We instead study class-level unpaired vision-language matching with known within-modality groups and an unordered candidate set of textual concepts, where visual groups and a set of candidate textual concepts are available but their cross-modal correspondence is unknown.
Supervision is confined to constructing class-level representations within each modality, with no paired image--text samples, cross-modal anchors, or verified correspondences provided to the matching procedure. When reliable cross-modal anchors are available, supervised alignment is often simpler and more practical.
Our goal is to determine whether independently trained encoders contain sufficient relational structure to recover correspondence without such anchors. This zero-anchor setting can arise when visual groups and a candidate textual ontology are collected or maintained independently, so that both sides are available but their correspondence has not been verified. In this regime, correspondence recovery can reduce the need for manually established cross-modal anchors and also serves as a diagnostic of structural compatibility between independently trained representation spaces. More broadly, it tests whether cross-modal correspondence can emerge from relational structure alone, without shared training data or explicit cross-modal supervision.
Recent work frames this problem as matching pairwise relationships between embeddings~\cite{schnaus2025blind}. Such formulations arise naturally in kernel alignment~\cite{cristianini2002kernel}, graph matching via quadratic assignment~\cite{vogelstein2015fast}, and structural optimal transport, most notably the Gromov-Wasserstein ($\mathcal{GW}$) distance~\cite{peyre2016gromov,xu2019gromov}.

\textbf{Limitations of value-based relational alignment.}
Blind matching compares the within-modality similarity kernels
$K_x$ and $K_y$ and searches for the class permutation $\pi$
under which their relational structures agree best (Section~\ref{sec:prelim}).
Existing relational alignment methods already operate at the level of pairwise relations rather than raw embeddings, but they still compare the numerical entries of the two kernels directly~\cite{moschella2022relative, schnaus2025blind, norelli2023asif}. Even when vision and language encoders learn similar embeddings, the numerical entries of their within-modality kernels need not be directly comparable across modalities. The two modalities may agree on the ordering of concept similarities, while still assigning different kernel values to the same relations. Consequently, alignment objectives that rely directly on kernel-entry comparison can become weakly discriminative~\cite{kornblith2019similarity,raghu2017svcca}.

This numerical mismatch weakens the discriminative value of direct kernel-entry comparison. At the same time, recent evidence suggests that cross-modal agreement can be more stable in relative and local relational structure than in absolute similarity values ~\cite{groger2026revisiting}. These observations raise a key question: can cross-modal correspondence be recovered from relational order when the numerical kernel values are not directly comparable across modalities?

\textbf{Rank-based relational alignment.}
To address this issue, we introduce an rank-based formulation of relational alignment. Instead of matching kernels based on raw values, we align them by comparing their rank-based transformation, which uses the ordering induced by these entries. The goal is therefore not to compare \(K_x(i,j)\) and \(K_y(\pi(i),\pi(j))\) as numbers, but to ask whether the two modalities induce consistent relational order over concept pairs.
Concretely, we replace each off-diagonal entry of a within-modality kernel with its rank among the off-diagonal entries and evaluate the same matching objective on the transformed kernels.
The correspondence problem itself remains unchanged, while the matching signal is determined by relational order rather than raw kernel values.

\textbf{Contributions.}
We study class-level unpaired vision-language matching with known
categories. Class-level representations are available
within each modality, while their cross-modal correspondence is unknown.
Our contributions are summarized as follows:
\begin{itemize}
    \item \textbf{Solver-agnostic ordinal relational representation.}
    We introduce a rank-based representation that plugs into existing blind-matching objectives and QAP solvers without changing the correspondence space or optimization procedure. RankAlign removes arbitrary strictly increasing reparameterizations of similarity values while retaining the ordinal structure used for matching.
    \item \textbf{Theory of ordinal correspondence recovery.}
    We formalize cross-modal agreement up to modality-specific monotone reparameterizations and show that the resulting ordinal representation is invariant to such distortions. We further provide exact-recovery and stability guarantees that separate cross-modal ordinal disagreement from the intrinsic distinguishability of class relational profiles.
    \item \textbf{Empirical validation.}
    Experiments across diverse vision and language encoders and datasets demonstrate that RankAlign achieves more stable and reliable cross-modal correspondence alignment.
\end{itemize}

\section{Method}

\begin{figure}[!tbp]
\centering
\includegraphics[width=0.95\linewidth]{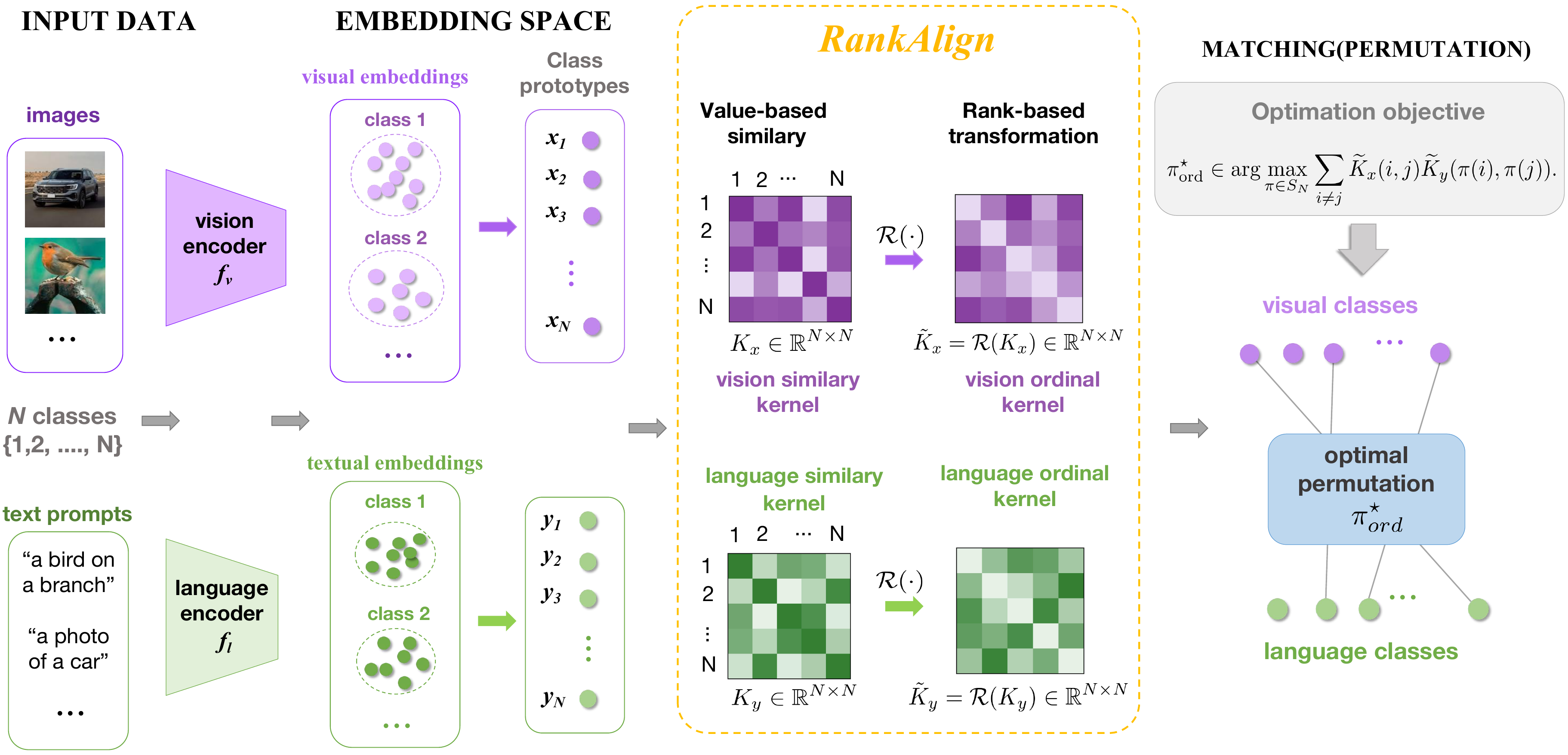}
\caption{
Visual-language alignment pipeline. The RankAlign module 
provides the core ordinal representation of each within-modality similarity kernel. These representations are refined using local neighbor order before permutation matching, without changing the matching space or solver.
}
\label{fig:RankAlign_framework}
\end{figure}

Figure~\ref{fig:RankAlign_framework} summarizes our visual-language alignment pipeline, which illustrates its key steps: \emph{i}) learning visual and textual embeddings and computing the corresponding prototypes; \emph{ii}) RankAlign transforms the visual and textual similarity kernels into rank-based similarity kernels; and \emph{iii}) permutation matching for cross-modal alignment. In the following sections, we will describe each component in detail.


\subsection{Preliminary}
\label{sec:prelim}

We adopt the blind-matching setting of \citet{schnaus2025blind}.
Let $f_v$ and $f_l$ denote frozen visual and language encoders, respectively. We consider $N$ semantic classes shared across the two modalities.   We construct class-level prototypes as
\begin{equation}
x_i =\frac{1}{|\mathcal I_i|}\sum_{I\in\mathcal I_i} f_v(I),
\qquad
y_i =\frac{1}{|\mathcal T_i|}\sum_{T\in\mathcal T_i} f_l(T).
\end{equation}
Then, we center the prototypes independently within each modality and renormalize them before constructing the relational kernels
\begin{equation}
\hat x_i=
\frac{x_i-\bar x}{\|x_i-\bar x\|},
\qquad
\bar x=\frac1N\sum_{j=1}^N x_j,
\end{equation}
and analogously for the language prototypes. This operation uses no cross-modal correspondence information and is permutation equivariant. 

For notational simplicity, we use $x_i$ and $y_i$ below to denote the resulting normalized prototypes. The within-modality similarity kernels are
\begin{equation}
K_x(i,j)=\langle x_i,x_j\rangle,
\qquad
K_y(i,j)=\langle y_i,y_j\rangle.
\end{equation}
A candidate cross-modal correspondence is represented by a permutation $\pi\in S_N$. Given a distortion function
$\ell:\mathbb R\times\mathbb R\rightarrow\mathbb R$, define
\begin{equation}
D_\ell(K_x,K_y;\pi)=\sum_{i\neq j}\ell\!\left(K_x(i,j),K_y(\pi(i),\pi(j))
\right).
\end{equation}
Blind relational matching seeks
\begin{equation}
\pi^\star \in \arg\min_{\pi\in S_N} D_\ell(K_x,K_y;\pi).
\end{equation}

Equivalently, we use the agreement score
\begin{equation}
\label{eq:agreement}
f(\pi)=\sum_{i\neq j} K_x(i,j) K_y(\pi(i),\pi(j)).
\end{equation}

\begin{lemma}[Distortion and agreement]
\label{lem:distortion}
For $\ell(a,b)=-ab$, minimizing $D_\ell$ is equivalent to maximizing $f$. For the squared distortion $\ell(a,b)=(a-b)^2$,
\begin{equation}
D_\ell(K_x,K_y;\pi)=C_x+C_y-2f(\pi),
\end{equation}
where $C_x$ and $C_y$ do not depend on $\pi$.
Hence both formulations have identical optimizers.
\end{lemma}
Hence, the objective function can be written as $\pi^\star \in \arg\max_{\pi\in S_N} f(K_x,K_y;\pi).$
And we use the maximization form throughout the remainder of the paper. The resulting problem is a quadratic assignment problem (QAP). A proof of Lemma~\ref{lem:distortion} is given in Appendix~\ref{app:representation}.

\subsection{Limitations of value-based relational alignment}
\label{sec:motivation}

The objective in Eq.~\ref{eq:agreement} compares numerical kernel values directly.
This implicitly assumes that visual and language similarities are
sufficiently calibrated across modalities. Such calibration is not
guaranteed for independently trained encoders.

To formalize this distinction, suppose both modalities encode a
shared latent relational structure $S=\{s_{ij}\}$ through
modality-specific strictly increasing maps,
\begin{equation}
\label{eq:latent-model}
K_x(i,j)=\phi_x(s_{ij}),
\qquad
K_y(\pi^\star(i),\pi^\star(j))=\phi_y(s_{ij}),
\end{equation}
where $\pi^\star$ denotes the ground-truth correspondence.

Even under the correct correspondence, the numerical similarities in the two modalities may differ substantially whenever $\phi_x$ and $\phi_y$ have different parameterizations. Consequently, a value-based objective can penalize a correct correspondence despite the two modalities encoding the same underlying relational structure.

By contrast, strictly increasing transformations preserve relational order. For any two unordered class pairs $e,e'$,
\begin{equation}
K_x(e)<K_x(e')
\quad\Longleftrightarrow\quad
K_y(\pi^\star(e))<K_y(\pi^\star(e')).
\end{equation}
Thus, the two modalities may share the same ordinal relational
geometry without sharing the same numerical parameterization.
This suggests representing relational structure through order rather than absolute similarity values, thereby making the matching objective insensitive to modality-specific monotone reparameterizations.

\subsection{Rank-based relational alignment}
\label{sec:RankAlign}

Because the similarity kernels are symmetric, we rank unordered class
pairs. Let
\begin{equation}
\mathcal E=\{\{i,j\}:1\leq i<j\leq N\},
\qquad
M=|\mathcal E|=\frac{N(N-1)}{2}.
\end{equation}

For a symmetric kernel $K$, we define the empirical cumulative
distribution function of its off-diagonal similarities as
\begin{equation}
F_K(t)=\frac{1}{M}\sum_{\{i,j\}\in\mathcal E}\mathbf 1[K(i,j)\leq t],
\end{equation}
and the corresponding rank transformation as
\begin{equation}
R(K)(i,j)=
\begin{cases}
F_K(K(i,j)), & i\neq j,\\
0, & i=j.
\end{cases}
\label{eq12}
\end{equation}

The transformation replaces the numerical spacing of pairwise similarities by their relative positions within each modality. If all $M$ unordered off-diagonal similarities are distinct, the transformed values are $1/M,\ldots,1$. If ties are present in the original kernel, they remain tied after the transformation. Rank transformation preserves the available relational ordering rather than creating additional relational information. 

\begin{lemma}[Ordinal invariance]
\label{lem:rank_invariance}
For every strictly increasing function
$\phi:\mathbb R\rightarrow\mathbb R$,
\begin{equation}
R(\phi\circ K)=R(K).
\label{eq:rank-invariance}
\end{equation}
Moreover, for every permutation $\sigma\in S_N$,
\begin{equation}
R(K\circ\sigma)=R(K)\circ\sigma,
\label{eq:rank-equivariance}
\end{equation}
where $(K\circ\sigma)(i,j)=K(\sigma(i),\sigma(j))$.
\end{lemma}

Lemma~\ref{lem:rank_invariance} establishes the central invariance of the rank representation. Modality-specific strictly monotone reparameterizations of similarity values are removed without changing the underlying correspondence structure. The transformation also preserves every strict pairwise ordering induced by the original kernel. The proof is provided in Appendix~\ref{app:representation}.

Replacing the kernels in Eq.~\ref{eq:agreement} by their rank
representations gives the ordinal matching objective
\begin{equation}
f_{\mathrm{ord}}(\pi)
=
\sum_{i\neq j}
\widetilde K_x(i,j)
\widetilde K_y(\pi(i),\pi(j)),
\qquad
\widetilde K_x=R(K_x),
\quad
\widetilde K_y=R(K_y).
\label{eq:ordinal-objective}
\end{equation}
The correspondence problem itself is unchanged. RankAlign changes only the relational representation supplied to the matching objective. The recovery analysis below focuses on the core rank representation $R(K)$, while the full ordinal refinement used in the experiments is described in Appendix~\ref{app:representation}.

\subsection{Recovery guarantees}
\label{sec:recovery}

We next give conditions under which ordinal relational matching
recovers the ground-truth correspondence. We first establish exact
recovery for the basic rank objective in Eq.~\ref{eq:ordinal-objective}, which isolates the contribution of
ordinal information, and then study stability under partial ordinal disagreement.

\begin{assumption}[Strict ordinal consistency]
\label{ass:ordinal}
Let $\pi^\star$ denote the ground-truth correspondence. Assume that the unordered off-diagonal values of $K_x$ are distinct and that the two modalities induce the same strict ordering over unordered class pairs under $\pi^\star$. 
For any distinct $e,e'\in\mathcal E$,
\begin{equation}
K_x(e)<K_x(e')
\quad\Longleftrightarrow\quad
K_y(\pi^\star(e))<K_y(\pi^\star(e')).
\label{eq:ordinal-consistency}
\end{equation}
\end{assumption}

Assumption~\ref{ass:ordinal} requires agreement in relational order, rather than agreement in numerical similarity values. The noise-free latent model of Eq.~\ref{eq:latent-model} satisfies this assumption whenever the latent similarities $s_{ij}$ are distinct, independently of the strictly increasing modality-specific maps $\phi_x$ and $\phi_y$.

\begin{proposition}[Exact recovery]
\label{prop:exact}
Under Assumption~\ref{ass:ordinal} and $N\geq3$,
$\pi^\star$ is the unique maximizer of
$f_{\mathrm{ord}}$ over $S_N$. Moreover, for every
$\pi\neq\pi^\star$,
\begin{equation}
f_{\mathrm{ord}}(\pi^\star)-f_{\mathrm{ord}}(\pi)
\geq
\frac{2}{M^2},
\qquad
M=\frac{N(N-1)}{2},
\label{eq:rank-margin}
\end{equation}
when the objective is summed over ordered off-diagonal pairs.
Equivalently, the corresponding margin over unordered pairs is at
least $1/M^2$.
\end{proposition}

Proposition~\ref{prop:exact} shows that strict ordinal consistency is sufficient for unique correspondence recovery. The guarantee depends only on the relational ordering and the problem size, rather than on the numerical spacing of either kernel. The complete proof is given in Appendix~\ref{app:recovery}.


Strict ordinal consistency is an idealized condition. In practice,
some relational orders may disagree across modalities. We therefore state the stability result in a representation-independent form. Let $Z(K)$ denote any permutation-equivariant ordinal representation, with
$Z_x=Z(K_x)$ and $Z_y=Z(K_y)$. We define the cross-modal residual as
\begin{equation}
E_Z = Z_y\circ\pi^\star - Z_x,
\label{eq:EZ}
\end{equation}
and the structural discrepancy induced by a permutation $\tau$ as
\begin{equation}
d_Z(\tau)
=
\|Z_x-Z_x\circ\tau\|_F.
\label{eq:dZ}
\end{equation}

\begin{proposition}[Stability under partial ordinal disagreement]
\label{prop:stability}
For any candidate permutation $\pi$, let
$\tau=(\pi^\star)^{-1}\circ\pi$. Then
\begin{equation}
f_Z(\pi^\star)-f_Z(\pi)
\geq
d_Z(\tau)
\left(
\frac{1}{2}d_Z(\tau)-\|E_Z\|_F
\right),
\label{eq:stability}
\end{equation}
where
$
f_Z(\pi)=\sum_{i\neq j}
Z_x(i,j)Z_y(\pi(i),\pi(j)).
$
Consequently, every maximizer $\widehat\pi$ satisfies
\begin{equation}
d_Z(\widehat\tau)
\leq
2\|E_Z\|_F,
\qquad
\widehat\tau =(\pi^\star)^{-1}\circ\widehat\pi.
\label{eq:stability-max}
\end{equation}
\end{proposition}

The result separates two sources of difficulty. The residual
$\|E_Z\|_F$ measures cross-modal ordinal disagreement, while
$d_Z(\tau)$ measures how distinguishable the relational structure is under an incorrect relabeling. Further specialization to the practical RankAlign representation, together with bounds on mismatched classes and approximate solver outputs, is provided in
Appendix~\ref{app:recovery}.


\subsection{Computational complexity}
\label{sec:complexity}

We finally analyze the additional computational cost introduced by RankAlign. Constructing a dense similarity kernel from $N$ class prototypes of dimension $d$ requires $O(N^2d)$ time and $O(N^2)$ memory. The kernel-level rank transformation sorts $M=N(N-1)/2=O(N^2)$ pairwise similarities, while the anchor-wise transformation ranks $N$ rows of length $N-1$. Both operations require $O(N^2\log N)$ time. The score transformation, symmetrization, and standardization are all $O(N^2)$.

Consequently, excluding feature extraction, RankAlign representation can be constructed in $O(N^2d+N^2\log N)$ time and $O(N^2+Nd)$ memory. Since the $O(N^2d)$ kernel construction is already required by value-based relational matching, the additional asymptotic preprocessing overhead of RankAlign is $O(N^2\log N)$ time and $O(N^2)$ memory. RankAlign does not change the subsequent correspondence problem. If $C_{\mathcal S}(N)$ denotes the running time of a chosen QAP solver $\mathcal S$, the complete matching pipeline requires $C_{\mathcal S}(N)+O(N^2d+N^2\log N)$ time. Further details are provided in Appendix~\ref{app:complexity}.

\section{Experiments}
We measured performance using matching accuracy, defined as the fraction of correctly recovered class correspondences. A matching
accuracy of $1.00$ corresponds to exact recovery of the ground-truth class permutation.
We compared against BlindMatch~\cite{schnaus2025blind} using both Gromov--Wasserstein (($\mathcal{GW}$)) and Mutual kNN. RankAlign uses the same permutation solvers as the baselines but replaces the raw within-modality similarity kernels with the ordinal representation defined in Section~\ref{sec:RankAlign} before matching.
To isolate the contribution of this representation, we additionally compare against z-score normalization, quantile normalization, and row-wise ranking under an identical protocol. 
The controlled ablation in Appendix~\ref{app:kernel_ablation} shows that the gains cannot be explained by generic normalization or local ranking alone. All experiments were conducted on a single NVIDIA RTX A6000 GPU with 48GB memory. 


\textbf{Pre-trained models}. We evaluated RankAlign using a variety of {frozen} pre-trained visual and textual encoders to cover diverse representation geometries and supervision paradigms, including DINO/DINOv2~\cite{caron2021dino,oquab2024dinov2}, DeiT~\cite{touvron2021training}, and vision-language models such as CLIP~\cite{radford2021clip}, SigLIP~\cite{zhai2023sigmoid}, and SigLIP2~\cite{tschannen2025siglip}. For language side, we primarily used SentenceTransformers models~\cite{reimers2019sentence}.

\textbf{Prompt construction.}
Given an image classification dataset with $N$ visual groups, we form class-level prototypes independently in the two modalities. On the visual side, each image is embedded with the frozen visual encoder and the embeddings are averaged within each visual group. On the language side, we construct an unordered candidate set of $N$ textual concepts. For each candidate concept, we generate a small collection of semantically equivalent textual descriptions, embed them with the frozen language encoder, and average them to obtain a textual prototype. The correspondence between the visual groups and textual prototypes is withheld from the matching algorithm. The algorithm receives only the
two unordered sets of class-level prototypes and recovers their
cross-modal permutation from within-modality relational structure. 

\textbf{Dataset}
We conducted our experiments on CIFAR-100 \cite{krizhevsky2009learning}, ImageNet-100 \cite{russakovsky2015imagenet}, the medical imaging dataset CheXpert \cite{irvin2019chexpert}, and a pseudo-class dataset derived from the LAION-Aesthetic subset \cite{schuhmann2022laion}. More details are provided in Appendix~\ref{app:data}.

\subsection{Computational Cost}
\label{sec:results}

Because encoders were frozen, no end-to-end training was required. RankAlign introduces only an ordinal preprocessing step before permutation matching. As analyzed in Section~\ref{sec:complexity}, constructing the complete RankAlign representation requires $O(N^2\log N)$ additional preprocessing time and $O(N^2)$ memory beyond the similarity-kernel construction. The transformed representations remain $N\times N$, so the overall computational cost is primarily determined by the QAP solver.

\begin{table}[htbp]
\centering
\caption{Runtime and matching accuracy on CIFAR-100 across different problem sizes using CLIP as the vision encoder and Sentence-T as the language encoder. Results are reported as mean $\pm$ standard deviation over five random seeds. $^\ast$ indicates that the solver reaches the 300s time limit. Actual wall-clock runtime may be slightly higher due to termination and post-processing overhead.}
\label{tab:runtime}
\setlength{\tabcolsep}{10.0pt}
\renewcommand{\arraystretch}{1.02}
\small

\begin{tabular}{llccc}
\toprule
Solver & Method & $N=50$ & $N=70$ & $N=100$ \\
\midrule

\multicolumn{5}{c}{Matching accuracy} \\
\midrule

\multirow{2}{*}{FAQ}
& \textsc{BlindMatch}
& 0.26 $\pm$ \text{\scriptsize 0.10}
& 0.14 $\pm$ \text{\scriptsize 0.09}
& 0.04 $\pm$ \text{\scriptsize 0.02} \\
& \textsc{RankAlign}
& 0.98 $\pm$ \text{\scriptsize 0.02}
& 0.79 $\pm$ \text{\scriptsize 0.05}
& 0.66 $\pm$ \text{\scriptsize 0.21} \\

\addlinespace[2pt]

\multirow{2}{*}{OT}
& \textsc{BlindMatch}
& 0.04 $\pm$ \text{\scriptsize 0.01}
& 0.02 $\pm$ \text{\scriptsize 0.02}
& 0.01 $\pm$ \text{\scriptsize 0.01} \\
& \textsc{RankAlign}
& 0.98 $\pm$ \text{\scriptsize 0.02}
& 0.79 $\pm$ \text{\scriptsize 0.05}
& 0.69 $\pm$ \text{\scriptsize 0.18} \\

\addlinespace[2pt]

\multirow{2}{*}{FHG}
& \textsc{BlindMatch}
& 0.77 $\pm$ \text{\scriptsize 0.02}
& 0.42 $\pm$ \text{\scriptsize 0.07}
& 0.15 $\pm$ \text{\scriptsize 0.03} \\
& \textsc{RankAlign}
& 0.98 $\pm$ \text{\scriptsize 0.02}
& 0.91 $\pm$ \text{\scriptsize 0.04}
& 0.79 $\pm$ \text{\scriptsize 0.02} \\

\midrule
\multicolumn{5}{c}{Runtime (s)} \\
\midrule

\multirow{2}{*}{FAQ}
& \textsc{BlindMatch}
& 0.54 $\pm$ \text{\scriptsize 0.18}
& 0.70 $\pm$ \text{\scriptsize 0.12}
& 1.44 $\pm$ \text{\scriptsize 0.21} \\
& \textsc{RankAlign}
& 0.36 $\pm$ \text{\scriptsize 0.02}
& 1.10 $\pm$ \text{\scriptsize 0.03}
& 2.70 $\pm$ \text{\scriptsize 0.31} \\

\addlinespace[2pt]

\multirow{2}{*}{OT}
& \textsc{BlindMatch}
& 0.12 $\pm$ \text{\scriptsize 0.03}
& 0.28 $\pm$ \text{\scriptsize 0.01}
& 0.82 $\pm$ \text{\scriptsize 0.01} \\
& \textsc{RankAlign}
& 0.42 $\pm$ \text{\scriptsize 0.04}
& 1.20 $\pm$ \text{\scriptsize 0.03}
& 2.80 $\pm$ \text{\scriptsize 0.28} \\

\addlinespace[2pt]

\multirow{2}{*}{FHG}
& \textsc{BlindMatch}
& 300.00$^\ast$
& 300.00$^\ast$
& 300.00$^\ast$ \\
& \textsc{RankAlign}
& 300.00$^\ast$
& 300.00$^\ast$
& 300.00$^\ast$ \\

\bottomrule
\end{tabular}
\end{table}

Table~\ref{tab:runtime} reports runtime and matching accuracy on CIFAR-100 with respect to the problem size $N$. 
When the same solver is used, RankAlign introduces only a modest runtime overhead while providing substantially higher matching accuracy. The OT-based solver remains computationally efficient across all tested problem sizes. In contrast, FHG reaches the runtime limit at larger problem sizes in our implementation. These results indicate that the main scalability bottleneck comes from permutation optimization rather than the rank transformation itself.

\subsection{Scalability}

\begin{figure}[htbp]
\centering
\includegraphics[width=0.49\linewidth]{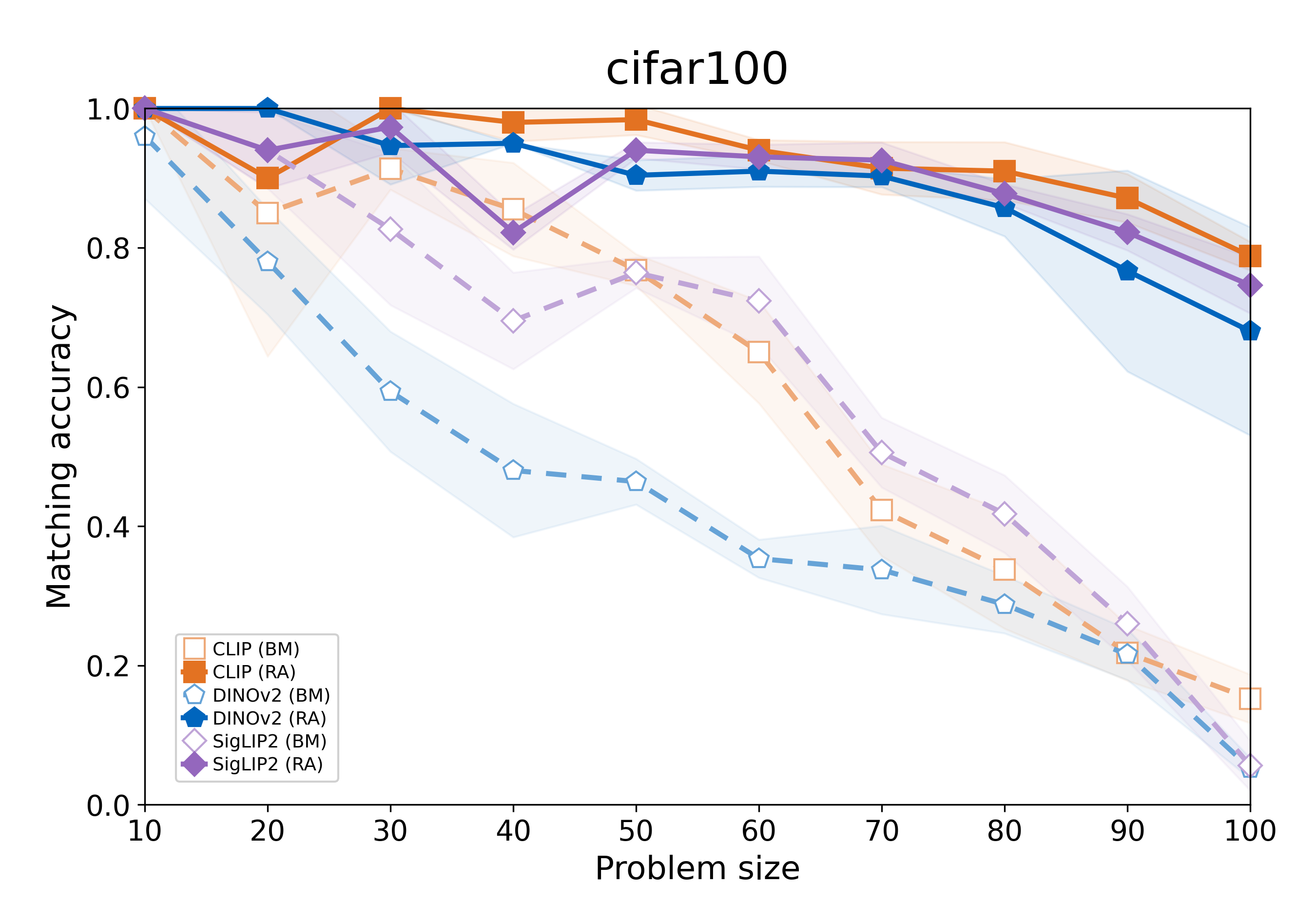}
\hfill
\includegraphics[width=0.49\linewidth]{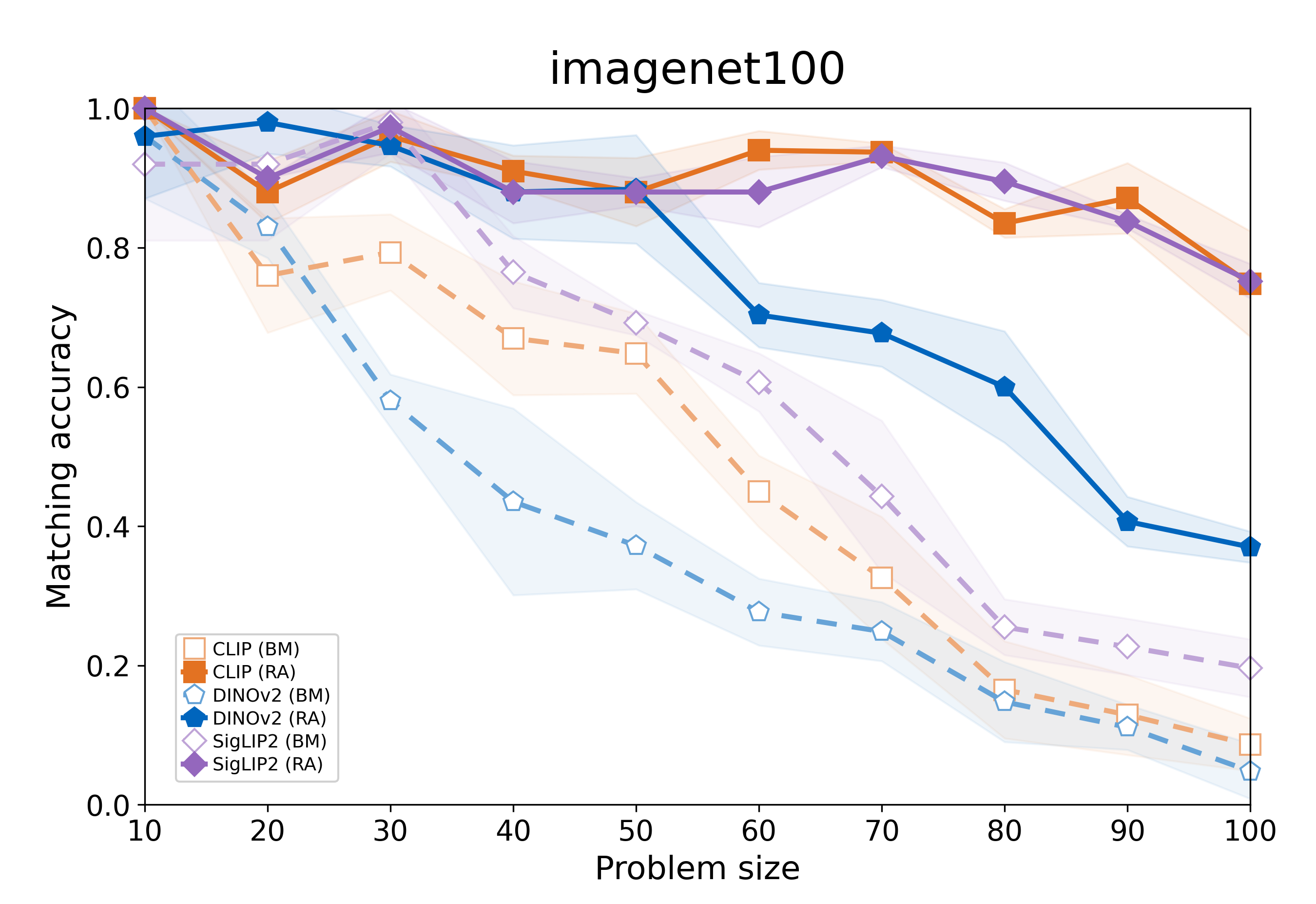}
\caption{
Matching accuracy as a function of problem size on CIFAR-100 and ImageNet-100. Each color corresponds to a vision backbone. For each backbone, the darker solid line denotes \textsc{RankAlign}, while the lighter dashed line denotes \textsc{BlindMatch} with $\mathcal{GW}$.}
\label{fig:largescale_matching}
\end{figure}

We tested scalability on CIFAR-100 and ImageNet-100 by increasing the problem size $N$, where larger $N$ requires finer-grained semantic discrimination and yields a more challenging matching problem.
Following~\cite{schnaus2025blind}, we selected class subsets using a p-dispersion strategy, which favors subsets whose relational structures are well aligned across modalities. Specifically, let \(S \subseteq \{1,\dots,L\}\) be a subset of classes with \(|S| = N\). 
We can get a subset score
$$A(S) = \sum_{i,j \in S} \ell\!\left(K_x(i,j), K_y(i,j)\right),$$
which measures the agreement between visual and language relational kernels restricted to \(S\). We then select subsets by approximately solving $S^\star \in \arg\max_{S: |S|=N} A(S),$ which is an instance of the p-dispersion-sum problem. 
This procedure yielded class subsets where cross-modal relational structure is more consistent, making non-trivial matching feasible. For each problem size \(N\), i.e., the number of classes to be matched, we constructed matching instances from the selected subsets and reported results across different visual encoders. 
We use the same solver of FHG~\cite{schnaus2025blind} for a fair comparison.
Fig.~\ref{fig:largescale_matching} shows that the matching problem
becomes progressively more challenging as $N$ increases, particularly at larger problem sizes. At smaller $N$, \textsc{RankAlign} reaches a matching accuracy of $1$
for several backbone settings. Since matching accuracy measures the fraction of correctly recovered class correspondences, a value of $1$ means that the recovered permutation exactly matches the ground-truth permutation, i.e., $\hat{\pi}=\pi^\star$. This should be interpreted as exact correspondence recovery for the given matching instance rather than as an indication of overfitting. This observation is consistent with Proposition~\ref{prop:exact}, which shows that under strict ordinal consistency, $\pi^\star$ is the unique maximizer of the ordinal matching objective.
As $N$ grows, \textsc{RankAlign} consistently outperforms \textsc{BlindMatch}($\mathcal{GW}$), with the advantage becoming increasingly pronounced at larger problem sizes. Most notably, at $N=100$, \textsc{RankAlign} retains approximately $30\%-60\%$
matching accuracy across the representative backbones on CIFAR-100 and ImageNet-100, while \textsc{BlindMatch}($\mathcal{GW}$) falls to near-chance levels. This yields an improvement of roughly $45\%-70\%$ at the largest evaluated problem size.

These observations are consistent with the recovery analysis in
Section~\ref{sec:recovery}. Ordinal consistency yields unique
recovery of the ground-truth permutation, while under partial
disagreement recovery depends on the balance between cross-modal
ordinal agreement and the distinguishability of class relational
profiles. The widening performance gap suggests that ordinal relational structure remains informative as correspondence recovery requires increasingly fine-grained discrimination among classes. For clarity, Fig.~\ref{fig:largescale_matching} reports representative results for CLIP, DINOv2, and SigLIP2, while results across all evaluated vision backbones are provided in
Appendix~\ref{app:largescale_full}. Additional comparisons in
Appendix~\ref{app:classical_alignment} show that classical Orthogonal Procrustes and vanilla Gromov--Wasserstein optimal transport remain substantially below RankAlign under the same representations and matching instances.

\subsection{Robustness to label corruption}
\label{app:robustness_label_flip}

\textbf{Setup.} We conducted this experiment on CIFAR-100 and ImageNet-100 with $N=100$ classes. For each corruption rate $r \in [0,1]$, we randomly assigned a fraction $r$ of image embeddings within each class to incorrect classes. Let $S_c$ be the set of samples originally assigned to class $c$, and let \(m_c=\lfloor r|S_c|\rfloor\) be the number of corrupted samples from class $c$. This produces corrupted labels $\tilde y_n$, while the language prototypes $\{t_c\}_{c=1}^C$ remain unchanged.
Using the corrupted labels, we recomputed the visual class centroids as
\[\tilde v_c=\frac{1}{|\{n:\tilde y_n=c\}|}\sum_{n:\tilde y_n=c} z_n,\]
where $z_n$ is the visual embedding of image $n$. We performed cross-modal matching between the corrupted visual prototypes $\{\tilde v_c\}_{c=1}^C$ and the clean language prototypes.

Let $\pi$ denote the recovered permutation. We report the class-level matching accuracy
\[\mathrm{Acc}_{\mathrm{match}}=\frac{1}{C}\sum_{c=1}^{C}\mathbf{1}\!\left[\pi(c)=c\right].\]
Each experiment was repeated over multiple random seeds and we report mean and standard deviation. Robustness is quantified by the relative drop from the clean setting:
\[\mathrm{Drop}(r)=\frac{\mathrm{Acc}(0)-\mathrm{Acc}(r)}{\mathrm{Acc}(0)} \times 100\%,\]
where $\mathrm{Acc}(r)$ denotes the mean matching accuracy at corruption rate $r$.

\begin{table}[htbp]
\centering
\caption{Robustness under visual-side label corruption on CIFAR-100 and ImageNet-100. We report matching accuracy (\%) as mean $\pm$ standard deviation over five random seeds and relative performance drop (\%) with respect to the clean setting.} 
\label{tab:label_flip_robustness_new}
\resizebox{0.95\linewidth}{!}{
\begin{tabular}{llcccccc}
\toprule
Dataset & Metric & 0.00 & 0.10 & 0.20 & 0.30 & 0.40 \\
\midrule
\multirow{2}{*}{CIFAR-100}
& Acc. (\%)  
& 78.7 $\pm$ 7.3
& 78.0 $\pm$ 5.2
& 77.6 $\pm$ 5.8
& 73.1 $\pm$ 15.9
& 68.9 $\pm$ 6.4
\\
& Drop (\%) 
& 0.0
& 0.9
& 1.4
& 7.1
& 12.5
\\
\midrule
\multirow{2}{*}{ImageNet-100}
& Acc. (\%)
& 74.4 $\pm$ 5.4
& 71.0 $\pm$ 3.9
& 71.4 $\pm$ 5.4
& 56.6 $\pm$ 12.0
& 41.5 $\pm$ 10.3
\\
& Drop (\%)
& 0.0
& 4.6
& 4.0
& 23.9
& 44.2
\\
\bottomrule
\end{tabular}}
\end{table}

Table~\ref{tab:label_flip_robustness_new} reports robustness under increasing visual-side label corruption on CIFAR-100 and ImageNet-100. As the corruption rate increases, matching accuracy decreases on both datasets, reflecting the impact of noisy class assignments on prototype quality. The degradation remains gradual across a wide range of corruption levels, indicating that our method maintains reasonable performance even under moderate noise. Similar trends are observed on both datasets, suggesting that the approach is consistently robust across different data distributions.

\subsection{LAION Pseudo-Class Matching}
\label{app:laion_matching}

We further evaluate RankAlign on the LAION pseudo-class dataset, which provides a noisy image--text setting without manually defined class annotations. We compare \textsc{RankAlign} against \textsc{BlindMatch}($\mathcal{GW}$) across different visual backbones.

\begin{table}[htbp]
\centering
\small
\caption{Matching accuracy on the LAION pseudo-class dataset. We compare \textsc{BlindMatch}($\mathcal{GW}$) with \textsc{RankAlign} across different visual backbones.}
\label{tab:laion_matching}
\setlength{\tabcolsep}{14pt}
\begin{tabular}{lcc}
\toprule
Vision model
& \textsc{BlindMatch}
& \textsc{RankAlign} \\
\midrule
DINOv2
& 0.59 $\pm$ {\scriptsize 0.29}
& 0.96 $\pm$ {\scriptsize 0.09} \\
DeiT
& 0.72 $\pm$ {\scriptsize 0.05}
& 0.76 $\pm$ {\scriptsize 0.09} \\
CLIP
& 0.77 $\pm$ {\scriptsize 0.13}
& 0.68 $\pm$ {\scriptsize 0.05} \\
SigLIP
& 0.54 $\pm$ {\scriptsize 0.13}
& 0.79 $\pm$ {\scriptsize 0.15} \\
SigLIP2
& 0.61 $\pm$ {\scriptsize 0.09}
& 0.74 $\pm$ {\scriptsize 0.11} \\
\bottomrule
\end{tabular}
\end{table}

As shown in Table~\ref{tab:laion_matching}, \textsc{RankAlign}
achieves stronger overall matching performance than
\textsc{BlindMatch}($\mathcal{GW}$) across the evaluated visual
backbones. The results indicate that rank-based relational alignment is generally more robust in this noisy pseudo-class setting, where class definitions are derived from weakly supervised image--text signals rather than manually curated labels. The consistent gains across different backbone families further suggest that the benefit of RankAlign is not tied to a specific visual encoder.

\subsection{Medical Domain Evaluation}

We also evaluate RankAlign on CheXpert, which introduces a substantial domain gap between medical images and language descriptions, together with subtle inter-class differences and class imbalance. Compared with natural-image benchmarks, CheXpert presents a more challenging
cross-modal matching setting because different pathological categories may exhibit highly similar visual patterns, while the corresponding language descriptions encode fine-grained clinical distinctions. Moreover, the uneven class distribution further perturbs class-level prototype estimation, making reliable correspondence recovery more difficult.
Table~\ref{tab:chexpert_results} reports matching accuracy across
different visual backbones. We compare \textsc{RankAlign} against
\textsc{BlindMatch}($\mathcal{GW}$) under the same evaluation setting. Across all evaluated backbones, \textsc{RankAlign} achieves substantially higher matching accuracy, indicating that the rank-based relational representation remains effective under pronounced domain shift and challenging medical-image semantics.

\begin{table}[htbp]
\centering
\small
\caption{Matching performance on the CheXpert subset. We report matching
accuracy across different vision backbones for
\textsc{BlindMatch}($\mathcal{GW}$) and \textsc{RankAlign}.}
\label{tab:chexpert_results}
\setlength{\tabcolsep}{16pt}
\begin{tabular}{lcc}
\toprule
Vision model
& \textsc{BlindMatch}
& \textsc{RankAlign} \\
\midrule
DeiT
& 0.08 $\pm$ {\scriptsize 0.01}
& 0.65 $\pm$ {\scriptsize 0.01} \\
DINOv2
& 0.06 $\pm$ {\scriptsize 0.01}
& 0.62 $\pm$ {\scriptsize 0.30} \\
CLIP
& 0.08 $\pm$ {\scriptsize 0.01}
& 0.63 $\pm$ {\scriptsize 0.26} \\

SigLIP
& 0.10 $\pm$ {\scriptsize 0.02}
& 0.45 $\pm$ {\scriptsize 0.22} \\

SigLIP2
& 0.08 $\pm$ {\scriptsize 0.00}
& 0.23 $\pm$ {\scriptsize 0.04} \\

\bottomrule
\end{tabular}
\end{table}

\section{Conclusions}
This work revisits class-level unpaired vision-language matching from an ordinal perspective and introduces RankAlign, a rank-based normalization strategy for relational correspondence. Rather than modifying the underlying permutation-matching framework, RankAlign changes the relational representation supplied to the matching objective. By emphasizing relational order over absolute similarity magnitudes, it provides a monotone-invariant signal for recovering cross-modal correspondences without cross-modal supervision. Beyond its empirical effectiveness, these findings offer strong support for the Platonic Representation Hypothesis, demonstrating that cross-modal convergence does not require rigid metric equivalence, but rather emerges naturally through shared relational topologies across independently trained representations.
A fundamental property of rank-based normalization lies in its complete invariance to strictly increasing reparameterizations, providing provable guarantees against modality-specific scale distortions, domain shifts, and monotonic noise. These theoretical properties translate directly into empirical robustness: across diverse frozen encoders and matching settings, RankAlign delivers stable class-level correspondence recovery even under severe corruption conditions. By formulating cross-modal correspondence in terms of relational order rather than absolute similarity magnitudes, the approach ensures invariant alignment without relying on cross-modal anchors or paired supervision.

\textbf{Limitations}
RankAlign relies on informative class-level prototypes in both modalities. When visual groups are estimated from noisy clustering rather than reliable class labels, prototype errors can distort the induced relational structure and reduce ordinal agreement, and our experiments suggest that prototype quality is a major source of degradation, although larger and more fine-grained matching problems may introduce additional difficulty. The permutation-based matching stage can also become computationally expensive as the number of classes grows. Moreover, RankAlign focuses on class-level zero-anchor correspondence and intentionally discards absolute similarity magnitudes, so instance-level, open-vocabulary, and hybrid ordinal--magnitude alignment remain directions for future work.

\textbf{Broader Impact}
RankAlign may reduce the need for manually verified cross-modal correspondences in settings where paired multimodal supervision is limited or expensive. Potential applications include representation analysis, class-level retrieval, and domain-specific multimodal matching. As with other alignment methods, unreliable prototypes or biases in pretrained models may lead to incorrect correspondences, so applications in sensitive domains should include appropriate validation.

\begin{ack}
We are grateful to Francis Engelmann for carefully reviewing the manuscript and providing valuable suggestions that helped improve the clarity of the paper.

\end{ack}

\bibliography{main}
\bibliographystyle{plainnat}

\newpage
\appendix


\section{Technical appendices and supplementary material}

In this appendix, we provide further theoretical results and detailed proofs in \S A. Additionally, we report supplementary empirical results and analyses in \S B.




\label{app:theory}

This appendix provides proofs and additional details for the
theoretical results in the main text. We first establish the relation between relational distortion and agreement and prove the invariance properties of the rank representation. We then describe the full ordinal refinement used in the experiments. Next, we prove the exact-recovery and stability results and connect them to this refined representation and approximate solver outputs. Finally, we provide additional computational complexity details.

\subsection{Properties of the ordinal representation}
\label{app:representation}

We first prove the relation between the distortion formulation and the
agreement score introduced in Section~2.1.

\paragraph{Proof of Lemma~\ref{lem:distortion}.}
For $\ell(a,b)=-ab$,
\[
D_\ell(K_x,K_y;\pi)
=
-\sum_{i\neq j}
K_x(i,j)K_y(\pi(i),\pi(j))
=
-f(\pi).
\]
Hence minimizing $D_\ell$ is equivalent to maximizing $f$.

For $\ell(a,b)=(a-b)^2$,
\begin{align}
D_\ell(K_x,K_y;\pi)
&=
\sum_{i\neq j}
\left(
K_x(i,j)-K_y(\pi(i),\pi(j))
\right)^2 \\
&=
\sum_{i\neq j}K_x(i,j)^2
+
\sum_{i\neq j}K_y(\pi(i),\pi(j))^2 \\
&\quad
-
2\sum_{i\neq j}
K_x(i,j)K_y(\pi(i),\pi(j)).
\end{align}
Because $(i,j)\mapsto(\pi(i),\pi(j))$ is a bijection over the
ordered off-diagonal pairs,
\[
\sum_{i\neq j}K_y(\pi(i),\pi(j))^2
=
\sum_{i\neq j}K_y(i,j)^2.
\]
The first two terms are therefore independent of $\pi$, which proves
the equivalence.

When the off-diagonal entries of both kernels have zero mean and unit
variance, define
\[
\rho(\pi)
=
\frac{1}{N(N-1)}
\sum_{i\neq j}
K_x(i,j)K_y(\pi(i),\pi(j)).
\]
Then
\[
f(\pi)=N(N-1)\rho(\pi),
\]
and
\[
D_\ell(K_x,K_y;\pi)
=
2N(N-1)(1-\rho(\pi)).
\]
Thus maximizing the agreement score is equivalent to maximizing the
correlation between the two relational structures.

\paragraph{Proof of Lemma~\ref{lem:rank_invariance}.}
Let $\phi$ be strictly increasing. For any observed kernel entries,
\[
\phi(K(k,l))\leq\phi(K(i,j))
\]
if and only if
\[
K(k,l)\leq K(i,j).
\]
Therefore
\[
F_{\phi\circ K}(\phi(K(i,j)))
=
F_K(K(i,j)),
\]
which implies
\[
R(\phi\circ K)=R(K).
\]

Now let $\sigma\in S_N$. Relabeling the classes induces a bijection on
the unordered-pair set $\mathcal E$ and therefore leaves the multiset
of off-diagonal kernel values unchanged. Hence
\[
R(K\circ\sigma)(i,j)
=
R(K)(\sigma(i),\sigma(j)),
\]
and therefore
\[
R(K\circ\sigma)=R(K)\circ\sigma.
\]

The rank transformation also preserves strict pairwise order. For
$e,e'\in\mathcal E$,
\[
K(e)<K(e')
\quad\Longleftrightarrow\quad
R(K)(e)<R(K)(e').
\]
If all $M=N(N-1)/2$ unordered off-diagonal values are distinct, the
rank values are $1/M,2/M,\ldots,1$. If ties occur, equal values remain
tied. Rank transformation therefore preserves the ordinal information
already present in the kernel rather than creating additional
relational distinctions.

The theoretical analysis in the main text is stated for the basic rank representation $R(K)$. In the experiments, we use a fixed ordinal refinement
$
T(K)=h(R(K)),
$
which combines kernel-level ranks with anchor-wise neighbor order.
For ranks computed over $m$ candidate relations, we use the monotone reweighting
\begin{equation}
g_m(u)=-\log\left(1-u+\frac{1}{2m}\right).
\label{eq:rank-score}
\end{equation}
Because $g_m$ is strictly increasing, it changes the relative weighting of rank positions without changing their ordering.
For each anchor $i$, define the within-row rank
\begin{equation}
R_{\mathrm{row}}(K)(i,j)
=
\frac{1}{N-1}
\sum_{l\neq i}
\mathbf{1}
\left[
R(K)(i,l)\leq R(K)(i,j)
\right],
\qquad i\neq j,
\label{eq:row-rank}
\end{equation}
with zero diagonal. Then let $\mathcal C(A)$ denote standardization over the off-diagonal
entries of a symmetric matrix $A$,
\begin{equation}
\mathcal C(A)(i,j)=\frac{A(i,j)-\mu_A}{s_A},
\qquad i\neq j,
\label{eq:offdiag-standardization}
\end{equation}
with zero diagonal, where $\mu_A$ and $s_A$ are the mean and standard deviation of the off-diagonal entries. We define
\begin{equation}
G(K)=\mathcal C\!\left(g_M(R(K))\right),
\label{eq:global-component}
\end{equation}
and
\begin{equation}
L(K)=\mathcal C\!\left(
\operatorname{sym}
g_{N-1}(R_{\mathrm{row}}(K))
\right),
\label{eq:local-component}
\end{equation}
where $\operatorname{sym}(A)=\frac{A+A^\top}{2}.$

The RankAlign representation is then $T(K)=\mathcal C\!\left(G(K)+L(K)\right).$

The two components are standardized before aggregation so that their relative contribution is not determined solely by numerical scale. Importantly, all operations in this refinement depend only on ordinal information. 
Consequently, $T(\phi\circ K)=T(K)$ for every strictly increasing $\phi$, and $T(K\circ\sigma)=T(K)\circ\sigma $
for every permutation $\sigma\in S_N$. Thus, the refined representation inherits the monotone invariance and
permutation equivariance of the basic rank representation without
reintroducing the original similarity magnitudes.

\subsection{Recovery guarantees and approximate optimization}
\label{app:recovery}

We first prove the exact-recovery result for the basic rank objective in Section~\ref{sec:recovery}.

\paragraph{Proof of Proposition~\ref{prop:exact}.}
Let $X=R(K_x)$, Under Assumption~\ref{ass:ordinal}, the two kernels induce the same strict ordering under $\pi^\star$. Hence
\[
R(K_y)\circ\pi^\star=X.
\]
For an arbitrary candidate permutation $\pi$, define
\[
\tau=(\pi^\star)^{-1}\circ\pi.
\]
Then
\[
R(K_y)\circ\pi=X\circ\tau.
\]
Therefore,
\begin{align}
f_{\mathrm{ord}}(\pi^\star)
-
f_{\mathrm{ord}}(\pi)
&=
\|X\|_F^2
-
\langle X,X\circ\tau\rangle \\
&=
\frac{1}{2}
\|X-X\circ\tau\|_F^2.
\end{align}
The right-hand side is non-negative for every $\pi$, so
$\pi^\star$ is a maximizer.

We next show uniqueness and the explicit margin. The permutation
$\tau$ induces a permutation of the unordered-pair set $\mathcal E$.
If every unordered pair were fixed, then for any $i$ and any two
distinct $j,k\neq i$,
\[
\tau(i)\in\{i,j\}\cap\{i,k\}=\{i\},
\]
so $\tau=\mathrm{id}$. Thus every non-identity class permutation with
$N\geq3$ changes at least one non-trivial cycle of unordered pairs.

Under Assumption~\ref{ass:ordinal}, the unordered rank values are
$1/M,\ldots,1$ and are pairwise distinct. A non-trivial cycle contains
at least two unordered pairs, and each changed rank differs by at least
$1/M$. Since every unordered pair contributes twice to the ordered
off-diagonal objective,
\[
\|X-X\circ\tau\|_F^2
\geq
\frac{4}{M^2}.
\]
Hence
\[
f_{\mathrm{ord}}(\pi^\star)
-
f_{\mathrm{ord}}(\pi)
\geq
\frac{2}{M^2}.
\]
Therefore $\pi^\star$ is the unique maximizer. If the objective is
instead written over unordered pairs only, each relation appears once,
giving the corresponding margin $1/M^2$.

The exact-recovery result above is stated for the basic rank
representation. Under Assumption~\ref{ass:ordinal}, the two modalities
differ only by a strictly increasing reparameterization under the
ground-truth correspondence. Since the refined representation $T(K)$ is monotone invariant and permutation equivariant,
\begin{equation}
T(K_y)\circ\pi^\star=T(K_x).
\label{eq:practical-consistency}
\end{equation}
Hence $\pi^\star$ also maximizes the practical RankAlign objective. It is unique whenever the transformed representation has no non-trivial permutation symmetry.
We now prove the stability result in the representation-independent form used in the main text. Let
\[
Z_x=Z(K_x),
\qquad
Z_y=Z(K_y),
\]
and define
$E_Z=Z_y\circ\pi^\star-Z_x.$
For
$
\tau=(\pi^\star)^{-1}\circ\pi,
$
let
$
d_Z(\tau)=\|Z_x-Z_x\circ\tau\|_F.
$

\paragraph{Proof of Proposition~\ref{prop:stability}.}
Since
$
Z_y\circ\pi^\star=Z_x+E_Z,
$
we have
$
Z_y\circ\pi=(Z_x+E_Z)\circ\tau.
$
Therefore,
\begin{align}
f_Z(\pi^\star)-f_Z(\pi)
&=
\langle Z_x,Z_x+E_Z\rangle
-
\langle Z_x,(Z_x+E_Z)\circ\tau\rangle \\
&=
\|Z_x\|_F^2
-
\langle Z_x,Z_x\circ\tau\rangle
+
\langle Z_x-Z_x\circ\tau^{-1},E_Z\rangle.
\end{align}
The first two terms satisfy
\[
\|Z_x\|_F^2
-
\langle Z_x,Z_x\circ\tau\rangle
=
\frac{1}{2}d_Z(\tau)^2.
\]
By Cauchy--Schwarz,
\[
\langle Z_x-Z_x\circ\tau^{-1},E_Z\rangle
\geq
-d_Z(\tau)\|E_Z\|_F.
\]
Hence
\[
f_Z(\pi^\star)-f_Z(\pi)
\geq
d_Z(\tau)
\left(
\frac{1}{2}d_Z(\tau)-\|E_Z\|_F
\right).
\]
If $\hat{\pi}$ maximizes $f_Z$, then we have
\[
f_Z(\pi^\star)-f_Z(\hat{\pi})\leq0,
\]
and
\[
d_Z(\hat{\tau})
\leq
2\|E_Z\|_F,
\qquad
\hat{\tau}
=
(\pi^\star)^{-1}\circ\hat{\pi}.
\]

The stability result also yields a bound on the number of mismatched
classes. Define
\[
\operatorname{err}(\pi)
=
|\{i:\pi(i)\neq\pi^\star(i)\}|
\]
and
\[
\kappa(Z_x)
=
\min_{\tau\neq\mathrm{id}}
\frac{
\|Z_x-Z_x\circ\tau\|_F^2
}{
|\{i:\tau(i)\neq i\}|
}.
\]
Then
\[
\kappa(Z_x)\operatorname{err}(\hat{\pi})
\leq
d_Z(\hat{\tau})^2,
\]
which gives
\begin{equation}
\operatorname{err}(\hat{\pi})
\leq
\frac{4\|E_Z\|_F^2}{\kappa(Z_x)}.
\label{eq:mismatch-bound}
\end{equation}

For the standardized representation $T$, define
\[
\rho^\star
=
\frac{
\langle T_x,T_y\circ\pi^\star\rangle
}{
N(N-1)
}.
\]
Because the off-diagonal entries are standardized to zero mean and
unit variance,
\[
\|E_T\|_F^2
=
2N(N-1)(1-\rho^\star).
\]
Substituting this expression into
Eq.~\ref{eq:mismatch-bound} gives
\[
\operatorname{err}(\hat{\pi})
\leq
\frac{
8N(N-1)(1-\rho^\star)
}{
\kappa(T_x)
}.
\]
This separates cross-modal ordinal disagreement, measured by
$\rho^\star$, from the intrinsic distinguishability of the relational
profiles, measured by $\kappa(T_x)$.

Finally, the previous result extends to an approximate QAP solver. Let
$\hat{\pi}$ denote an arbitrary solver output and define its objective
deficit relative to the ground truth as
\[
\eta
=
\max\{0,f_Z(\pi^\star)-f_Z(\hat{\pi})\}.
\]
From the stability inequality,
\[
d_Z(\hat{\tau})
\left(
\frac{1}{2}d_Z(\hat{\tau})-\|E_Z\|_F
\right)
\leq
\eta.
\]
Solving the resulting quadratic inequality gives
\[
d_Z(\hat{\tau})
\leq
\|E_Z\|_F
+
\sqrt{
\|E_Z\|_F^2+2\eta
}.
\]
Hence,
\[
\operatorname{err}(\hat{\pi})
\leq
\frac{
\left(
\|E_Z\|_F+
\sqrt{\|E_Z\|_F^2+2\eta}
\right)^2
}{
\kappa(Z_x)
}.
\]

The analysis therefore separates three sources of difficulty.
Cross-modal disagreement is captured by $\|E_Z\|_F$, structural
ambiguity by $\kappa(Z_x)$, and optimization error by $\eta$.

\subsection{Computational complexity}
\label{app:complexity}

We finally provide a detailed accounting of the computational cost of RankAlign. Let $N$ denote the number of classes and $d$ the embedding dimension. The number of unordered class pairs is
$$
M=\frac{N(N-1)}{2}.
$$

Prototype centering and normalization require $O(Nd)$ time.
Constructing each dense similarity kernel requires $O(N^2d)$ time and $O(N^2)$ memory. The kernel-level rank transformation sorts $M=O(N^2)$ values and requires $O(M\log M)=O(N^2\log N)$ time.

The anchor-wise transformation ranks $N$ rows of length $N-1$ and
also requires $O(N^2\log N)$ time. The score transformation, symmetrization, and standardization each
operate over at most $O(N^2)$ entries and therefore require
$O(N^2)$ time. Consequently, the complete practical RankAlign representation can be constructed in
$
O(N^2d+N^2\log N)
$
time and
$
O(N^2+Nd)
$
memory, excluding feature extraction.

The $O(N^2d)$ kernel construction is already required by conventional value-based relational matching. The additional asymptotic preprocessing overhead introduced by RankAlign is $O(N^2\log N)$ time and $O(N^2)$ memory. Let $C_{\mathcal S}(N)$ denote the running time of a chosen QAP solver $\mathcal S$. The complete pipeline requires $C_{\mathcal S}(N)+O(N^2d+N^2\log N)$ time.

The transformed representation remains an $N\times N$ matrix.
RankAlign therefore leaves the input structure of the QAP unchanged and can be combined with the same solver used by value-based relational matching. Evaluating the complete matching objective for one permutation requires $O(N^2)$ operations. If a local-search solver swaps two assignments, only terms involving the affected classes change, so the corresponding objective update involves $O(N)$ pairwise terms. We use an iterated local search (ILS) procedure to obtain a strong initial solution before applying the selected QAP solver. The ILS optimizes the rank-based objective and its solution is used to initialize FAQ, entropic OT, or FHG. The same initialization procedure is applied to BlindMatch for a fair comparison.

\section{Additional experiments}
In this section, we provide supplementary experimental results supporting the main paper.

\subsection{Datasets}
\label{app:data}

We evaluate RankAlign on diverse benchmarks spanning web-scale pseudo-label data, object recognition datasets at different semantic granularities, and a domain-specific medical imaging dataset.

\textbf{CIFAR-100 and ImageNet-100.}
To study scalability, we evaluated the matching accuracy as a function of the number of classes $N$ using subsets of CIFAR-100 \cite{krizhevsky2009learning} and ImageNet-100 \cite{russakovsky2015imagenet}. We varied $N$ across matching instances to assess performance under increasing task difficulty.

\textbf{LAION pseudo-class dataset.}    
We evaluated RankAlign on a pseudo-class dataset derived from the LAION-Aesthetic subset~\cite{schuhmann2022laion}. Since LAION lacks predefined class labels, we defined \(N=20\)  visual categories and assign pseudo-labels via caption matching. To ensure label quality, we classified each image using both its original caption (\texttt{TEXT}) and a \texttt{Moondream}-generated auxiliary caption, retaining only samples with a unique, consistent top-scoring assignment. The dataset contains approximately \(7{,}000\) samples, with \(200\)--\(500\) images per class. Visual and language prototypes are constructed by averaging embeddings within each pseudo-class and its associated captions, respectively.

\textbf{Medical imaging dataset.}
We evaluated RankAlign on CheXpert~\cite{irvin2019chexpert}, a large-scale chest X-ray benchmark with 220,000 radiographs. We extracted a 12-class single-label subset by retaining samples with one positive finding, totaling about 62,000 images. This setting is challenging due to subtle inter-class differences, class imbalance, and the medical image-text domain gap.


\subsection{Kernel Transformation Ablation}
\label{app:kernel_ablation}

To determine whether the gains of RankAlign can be explained by generic normalization or by a single form of ranking, we conduct a controlled ablation on CIFAR-100. All variants use identical visual and language prototypes, class subsets, similarity kernels, encoders, and permutation solvers. Only the relational representation supplied to the solver is changed.

We compare the raw kernel without transformation (none), z-score
normalization, row-wise ranking, quantile normalization, and the full RankAlign representation.  Z-score normalization removes global mean and scale differences while preserving the relative spacing of kernel values. Row-wise ranking preserves local neighbor order but does not define a single kernel-level ordinal ordering. Quantile normalization aligns marginal value distributions. 

\begin{table}[htbp]
\centering
\caption{Kernel transformation ablation on CIFAR-100. All variants use the same prototypes, class subsets, encoders, kernels, and permutation solver, only the kernel transformation differs.}
\label{tab:kernel_ablation}
\small
\setlength{\tabcolsep}{6pt}
\renewcommand{\arraystretch}{1.05}

\begin{tabular}{lcccc}
\toprule
Transformation
& $N=30$ & $N=50$ & $N=70$ & $N=100$ \\
\midrule
None                   & 0.03 & 0.01 & 0.00 & 0.00 \\
Z-score                & 0.03 & 0.02 & 0.01 & 0.00 \\
Row-wise ranking       & 0.32 & 0.14 & 0.09 & 0.06 \\
Quantile normalization & 0.06 & 0.07 & 0.04 & 0.00 \\
{RankAlign}            & 1.00 & 0.98 & 0.91 & 0.79 \\
\bottomrule
\end{tabular}
\end{table}

Table~\ref{tab:kernel_ablation} shows that the improvement cannot be explained by normalization alone. Z-score normalization provides almost no gain over the raw value-based kernel, consistent with the fact that affine normalization does not alter the optimizer of the inner-product matching objective. Quantile normalization also yields limited improvements. Row-wise ranking performs substantially better than value-based normalization, confirming that local ordinal information is useful.
However, its advantage decreases as the problem size grows. RankAlign consistently achieves the highest matching accuracy across all problem sizes, indicating that the combination of kernel-level ordinal normalization and anchor-wise neighborhood order is more effective than generic normalization or local ranking alone.

\subsection{Comparison with Classical Structural Alignment Methods}
\label{app:classical_alignment}

To examine whether simpler classical alignment methods can recover the same cross-modal correspondence without RankAlign's ordinal formulation, we compare against unsupervised Orthogonal Procrustes and standard entropic Gromov--Wasserstein optimal transport. All methods use the same class-level visual and language representations and are evaluated on the same CIFAR-100 matching instances, so the comparison isolates differences in the alignment objective rather than differences in the underlying encoders or prototypes.

\begin{table*}[htbp]
\centering
\small
\caption{Comparison with classical structural alignment methods on CIFAR-100. All methods use the same class-level representations and matching instances. We report matching accuracy across problem sizes.}
\label{tab:classical_alignment}
\setlength{\tabcolsep}{8pt}

\begin{tabular}{llcccc}
\toprule
Model & Method
& $N=30$
& $N=50$
& $N=70$
& $N=100$ \\
\midrule

\multirow{3}{*}{CLIP}
& Procrustes    & $0.01 \pm \text{\scriptsize 0.02}$ & $0.03 \pm \text{\scriptsize 0.01}$ & $0.02 \pm \text{\scriptsize 0.02}$ & $0.00 \pm \text{\scriptsize 0.00}$ \\
& Vanilla GW-OT & $0.03 \pm \text{\scriptsize 0.03}$ & $0.03 \pm \text{\scriptsize 0.02}$ & $0.03 \pm \text{\scriptsize 0.03}$ & $0.03 \pm \text{\scriptsize 0.00}$ \\
& RankAlign     & $1.00 \pm \text{\scriptsize 0.00}$ & $0.98 \pm \text{\scriptsize 0.02}$ & $0.91 \pm \text{\scriptsize 0.04}$ & $0.79 \pm \text{\scriptsize 0.02}$ \\
\midrule

\multirow{3}{*}{DeiT}
& Procrustes    & $0.04 \pm \text{\scriptsize 0.04}$ & $0.02 \pm \text{\scriptsize 0.01}$ & $0.02 \pm \text{\scriptsize 0.01}$ & $0.01 \pm \text{\scriptsize 0.00}$ \\
& Vanilla GW-OT & $0.07 \pm \text{\scriptsize 0.04}$ & $0.05 \pm \text{\scriptsize 0.03}$ & $0.07 \pm \text{\scriptsize 0.05}$ & $0.05 \pm \text{\scriptsize 0.00}$ \\
& RankAlign     & $0.93 \pm \text{\scriptsize 0.01}$ & $0.95 \pm \text{\scriptsize 0.05}$ & $0.72 \pm \text{\scriptsize 0.05}$ & $0.36 \pm \text{\scriptsize 0.16}$ \\
\midrule

\multirow{3}{*}{DINOv2}
& Procrustes    & $0.05 \pm \text{\scriptsize 0.04}$ & $0.02 \pm \text{\scriptsize 0.02}$ & $0.01 \pm \text{\scriptsize 0.01}$ & $0.04 \pm \text{\scriptsize 0.00}$ \\
& Vanilla GW-OT & $0.01 \pm \text{\scriptsize 0.02}$ & $0.01 \pm \text{\scriptsize 0.01}$ & $0.01 \pm \text{\scriptsize 0.01}$ & $0.00 \pm \text{\scriptsize 0.00}$ \\
& RankAlign     & $0.95 \pm \text{\scriptsize 0.06}$ & $0.90 \pm \text{\scriptsize 0.02}$ & $0.90 \pm \text{\scriptsize 0.04}$ & $0.68 \pm \text{\scriptsize 0.15}$ \\

\bottomrule
\end{tabular}
\end{table*}

Table~\ref{tab:classical_alignment} shows that neither classical baseline approaches the performance of RankAlign across the evaluated backbones and problem sizes. Both Orthogonal Procrustes and vanilla GW-OT remain close to chance as $N$ increases, whereas RankAlign retains substantially higher matching accuracy. Because all methods operate on the same class-level representations, this difference cannot be attributed to stronger encoders or higher-quality prototypes. These results indicate that a global linear alignment or direct matching of the original metric structures is insufficient to recover the cross-modal correspondence in these experiments, and RankAlign is not replacing a comparably accurate simpler baseline.

\subsection{Full larger-scale matching results}
\label{app:largescale_full}

In the main paper, we report representative larger-scale matching results for a subset of vision backbones to improve readability. We provide the full results across all evaluated backbones, including CLIP, DINOv2, DeiT, SigLIP, and SigLIP2.

\begin{figure}[htbp]
\centering
\includegraphics[width=0.49\linewidth]{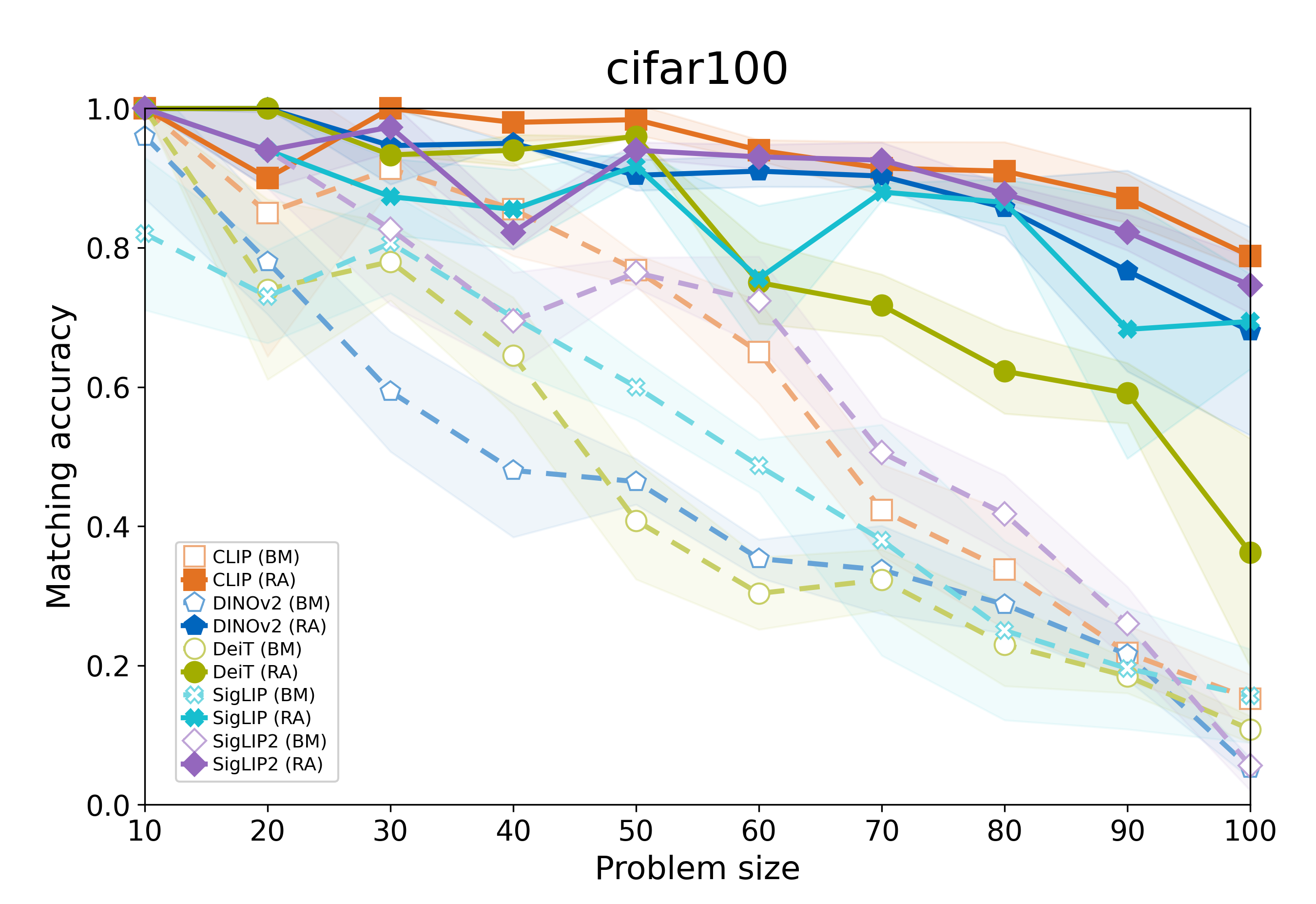}
\includegraphics[width=0.49\linewidth]{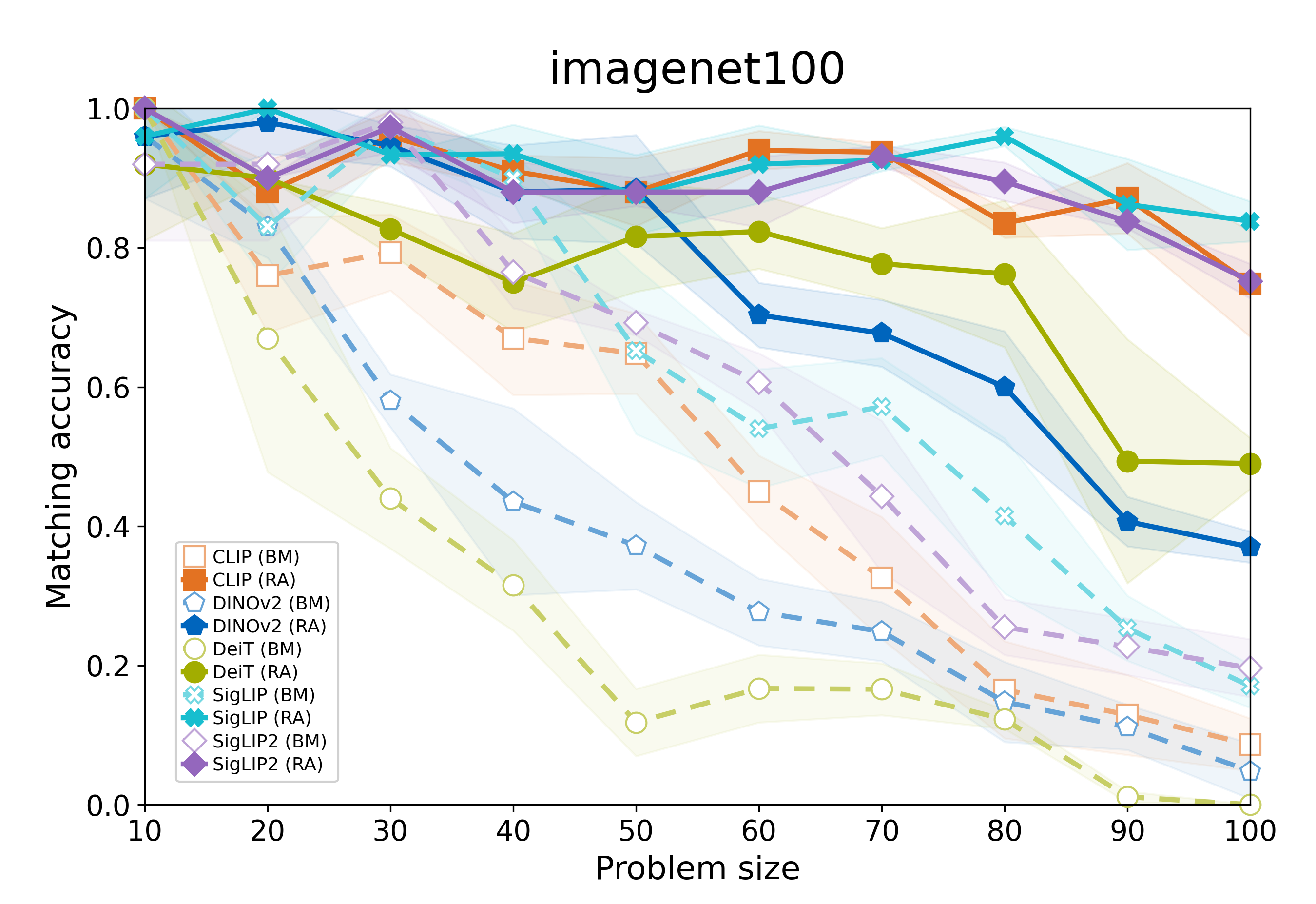}


\caption{
Full larger-scale matching results on CIFAR-100 and ImageNet-100 across all evaluated vision backbones.Each color corresponds to a vision backbone. The darker solid line denotes \textsc{RankAlign}, while the lighter dashed line denotes \textsc{BlindMatch} with the $\mathcal{GW}$.
}
\label{fig:largescale_matching_full}
\end{figure}

Figure~\ref{fig:largescale_matching_full} shows matching accuracy as a function of the problem size on CIFAR-100 and ImageNet-100.
The overall trend is consistent with the representative results in the main text, \textsc{RankAlign} consistently outperforms \textsc{BlindMatch} with the $\mathcal{GW}$ objective across different backbone families, and its performance degrades more gradually as the number of matched classes increases.
These results indicate that the improvement is not tied to a particular visual encoder, but holds across both vision-language and visual representation models.

\subsection{Existing Assets and Licenses}
This work uses existing public datasets and pretrained models, including LAION-Aesthetic, CIFAR-100, ImageNet-100, CheXpert, DINO/DINOv2, DeiT, CLIP, SigLIP/SigLIP2, and SentenceTransformers. We cite the original sources for these assets and use them only for research evaluation. We will follow the corresponding licenses and terms of use for each dataset and pretrained model, and will include detailed license information in the released code and supplemental documentation.

\end{document}